\documentclass[11pt]{article}

\usepackage[preprint]{acl}

\usepackage{times}
\usepackage{latexsym}
\usepackage{xspace}
\usepackage{amsmath}
\usepackage{amsfonts}
\usepackage{amssymb}
\usepackage{booktabs}
\usepackage{multirow}
\usepackage{colortbl}
\usepackage{xcolor}
\usepackage{makecell}
\usepackage{algorithm}
\usepackage{algpseudocode}
\usepackage{cleveref}
\usepackage{enumitem}

\usepackage[T1]{fontenc}

\usepackage[utf8]{inputenc}

\usepackage{microtype}

\usepackage{inconsolata}

\usepackage{graphicx}
\usepackage{subcaption}
\usepackage{tikz}
\usetikzlibrary{arrows.meta,calc}

\newcommand{\method}{\textsc{SAVE}\xspace}

\crefname{figure}{Figure}{Figures}
\crefname{table}{Table}{Tables}
\crefname{section}{Section}{Sections}
\crefname{equation}{Equation}{Equations}
\crefname{algorithm}{Algorithm}{Algorithms}

\title{The Flip Side of RLHF: On-Policy Feedback for Reward Model Self-Supervised Improvement}

\author{
  Xiaobo Wang\textsuperscript{1,4},
  Tong Wu\textsuperscript{4},
  Min Tang\textsuperscript{2},
  Jiaqi Li\textsuperscript{4},
  Qi Liu\textsuperscript{1,3*},
  Zilong Zheng\textsuperscript{4*}
  \\
  \textsuperscript{1}State Key Laboratory of Cognitive Intelligence, University of Science and Technology of China\\
  \textsuperscript{2}University of Science and Technology of China\\
  \textsuperscript{3}Institute of Artificial Intelligence, Hefei Comprehensive National Science Center\\
  \textsuperscript{4}State Key Laboratory of General Artificial Intelligence, BIGAI
  \\
}

\begin{document}
\maketitle
\begin{abstract}
Building strong reward models (RMs) for language model alignment is bottlenecked by the cost and difficulty of acquiring diverse and reliable preference data from human annotation or judge models. It is dramatically worse as the policy evolves beyond the static RM training. Therefore, we propose \textbf{\method} (\textbf{S}elf-supervised reward model improvement via \textbf{V}alue-\textbf{A}nchored On-policy fe\textbf{E}dback), a framework that grades on-policy responses as feedback by using the value function for on-policy RM training. \method naturally converts the reward-graded on-policy responses into supervision with a prompt-specific value head as an adaptive anchor. It computes RM advantages and filters ambiguous samples to update the RM via a contrastive objective. The effectiveness of \method for enhancing RM training is strongly validated through rigorous empirical evaluation across six diverse benchmarks. It achieves outperforming results across all datasets while maintaining consistent improvements across three RL algorithms (GRPO, RLOO, GSPO) and different policy backbones. 


\end{abstract}

\section{Introduction}

Large language models (LLMs) have demonstrated remarkable capabilities in a wide range of tasks, fundamentally reshaping natural language processing and artificial intelligence in a more general way \cite{touvron2023llama2, qwen2.5, qwen3technicalreport}. To align them with human preferences, modern post-training pipelines often use reinforcement learning from human feedback (RLHF) \cite{ouyang2022training}. In RLHF, a reward model (RM) trained on preference data provides the proxy objective for policy optimization, making RM quality central to aligned model performance.


Despite its central role, building a strong RM remains constrained by two bottlenecks. First, obtaining diverse, high-quality preference data is expensive: it either requires substantial human effort and careful quality control, or is distilled from stronger models which can be difficult to scale or iterate independently \cite{wang2024interpretable, SkyworkRewardV2}. Second, assigning reliable preference signals to such data is equally challenging. Human judgments are limited by annotation cost, whereas labels distilled from stronger external judge models \cite{lee2023rlaif, cui2024ultrafeedback} create dependence on an external oracle and ultimately upper-bound the RM by the teacher's capability. These limitations are amplified during RL training because the RM is trained on offline data while the policy continues to evolve. As the policy improves, its generation distribution drifts away from the static RM training set, leaving the RM increasingly miscalibrated in the regions visited most often by optimization \cite{casper2023open, coste2024reward}. This distributional mismatch induces a failure mode predicted by Goodhart's Law \cite{goodhart1984problems}: reward maximization amplifies errors in under-supervised regions, leading to reward hacking and over-optimization \cite{gao2023scaling}. Conventional remedies, such as collecting fresh human labels or re-querying an external judge, simply reproduce the original cost bottlenecks or perpetuate dependence on external supervision.



However, in standard RL, the value function can already serve as a natural reward baseline. By comparing a response’s reward with the prompt-level value, we can judge whether the response is relatively good or bad, thereby obtaining a supervised signal. Meanwhile, for the RM to keep refining its decision boundary, its training data must remain informative. Static policy responses soon become stale after the RM absorbs their supervision, as they no longer reveal the model’s current weaknesses. Therefore, the RM needs a data source that evolves with its capability. The continually updated RL policy naturally provides such data: as the policy improves, it generates fresh on-policy responses that target uncertain regions of the RM, supplying informative data for RM self-improvement.

This raises a natural question: \textit{can a reward model improve itself from on-policy responses generated during RL training, without additional human labels or an external judge?}

To answer this question, we propose \textbf{\method} (\textbf{S}elf-supervised reward model improvement via \textbf{V}alue-\textbf{A}nchored On-policy fe\textbf{E}dback), a general RM training framework that grades on-policy responses as self-supervised feedback for continuous reward model improvement. The key idea is to augment the RM with a prompt-specific value head that estimates the expected reward under the current sampling policy. This value estimate acts as an adaptive anchor for computing response-level RM advantages. We filter out ambiguous samples whose advantage magnitudes fall below a curriculum-driven threshold, partition the remaining responses into positive and negative advantage subsets, and use them as self-supervised feedback to update the reward model with a contrastive objective.


We further provide a theoretical interpretation of this framework. We show that value-anchored on-policy feedback can be formalized as a reward-model-centric minimax objective: the policy acts as an adaptive data generator that searches for challenging on-policy samples, while the RM minimizes its worst-case self-supervised ranking and calibration loss over the induced response distribution.

Empirically, extensive experiments validate the effectiveness of \method. On six reward model benchmarks (RewardBench, RewardBench 2, RM-Bench, PPE Preference, PPE Correctness, and JudgeBench), \method improves the average accuracy of the initial RM from 76.0 to 77.3, achieving the best scores on all six benchmarks. The improvements are consistent across three RL algorithms (GRPO, RLOO, and GSPO) and two policy backbones. Moreover, the improved RM further strengthens downstream RLHF policy performance: on AlpacaEval 2, the length-controlled win rate increases from 51.68\% to 54.24\%, and on Arena-Hard-v2.0, the win rate rises from 30.2\% to 33.9\%.

Our contributions are summarized as follows:
\begin{itemize}
    \item We propose \method, a general self-supervised framework that continuously improves reward models using on-policy feedback from RL training, without additional human annotation or external judges.
    \item We formulate \method as a reward-model-centric minimax problem, theoretically explaining why policy optimization naturally yields informative data for RM improvement.
    \item We empirically demonstrate consistent RM gains across six reward model benchmarks and show that the improved RM further enhances downstream RLHF policy performance.
\end{itemize}

\begin{figure*}[t]
    \centering
    \includegraphics[width=0.9\linewidth]{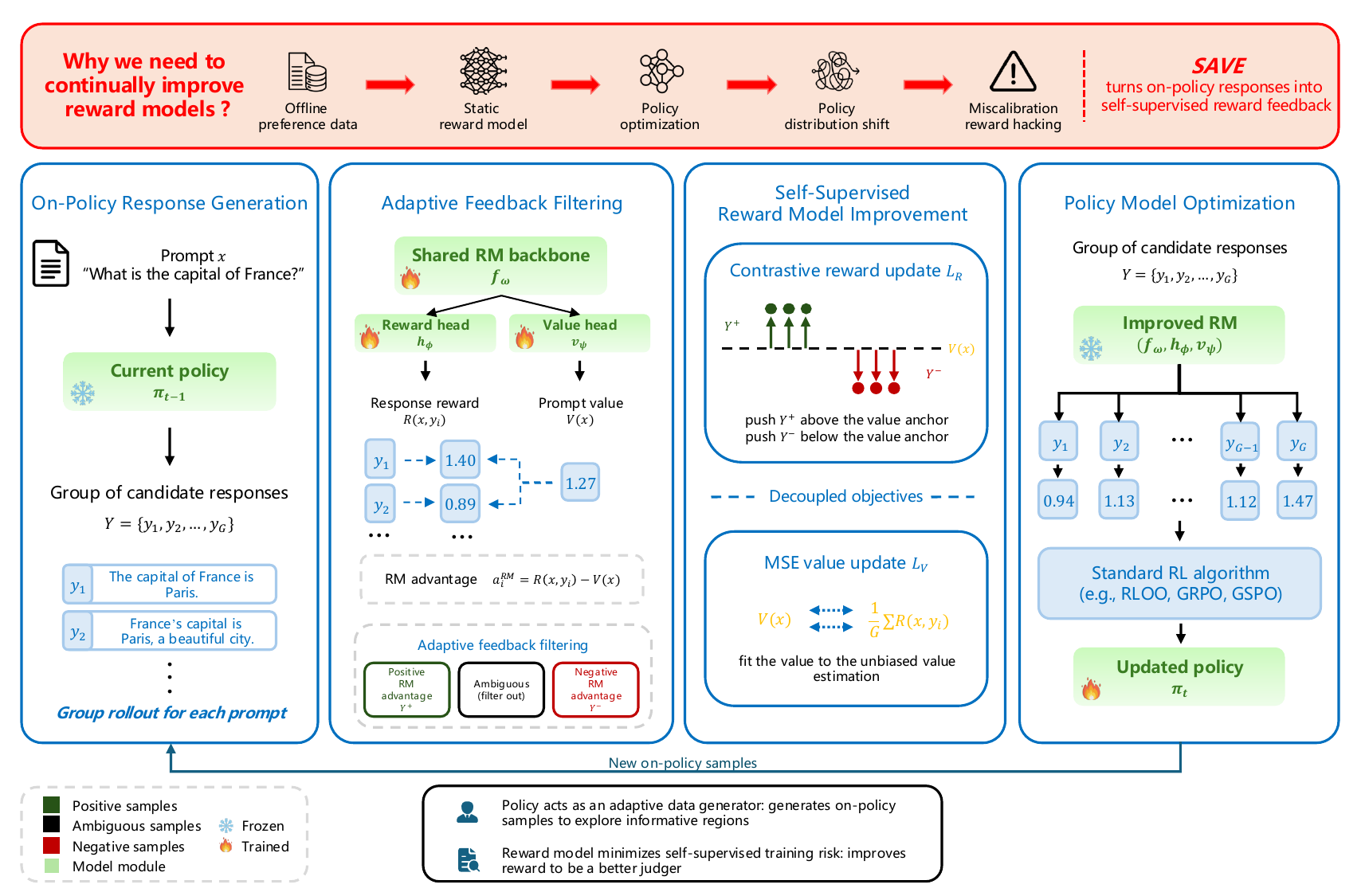}
    \caption{Overview of \method. At each training step $t$, the current policy samples an on-policy response group for each prompt. A value-anchored reward model computes response-level RM advantages, filters ambiguous samples, and partitions the retained responses into positive and negative feedback. The reward head is improved with a value-anchored contrastive objective, the value head is calibrated to the group mean reward, and the improved reward model supplies fresh rewards for policy model optimization.}
    \label{fig:savor_overview}
\end{figure*}

\section{Related Work}

\paragraph{Reward Modeling for Preference Alignment.}
Reward models (RMs) trained with the Bradley-Terry objective \cite{bradley1952rank} on pairwise preferences are the standard proxy reward in RLHF, but scalar RMs trained on fixed offline data suffer from over-optimization and reward hacking once the policy drifts off their training support \cite{gao2023scaling,casper2023open,coste2024reward}. Prior work improves RMs by scaling data and model capacity \cite{SkyworkRewardV2,yuan2024advancing}, adding robustness-oriented architectures \cite{wang2024interpretable,yang2024regularizing}, or introducing auxiliary calibration signals \cite{wang2025adaptive,nikulkov2026reward}.

\paragraph{Reinforcement Learning Algorithms for LLMs.}
Classical policy-gradient methods use value baselines to reduce variance \cite{williams1992simple,sutton2018reinforcement}, with PPO-style RLHF inheriting this design through advantage estimation \cite{schulman2016high,schulman2017proximal}. Recent critic-free methods such as RLOO \cite{ahmadian2024back}, GRPO \cite{shao2024deepseekmath}, REINFORCE++ \cite{hu2025reinforce}, DAPO \cite{yu2026dapo}, and GSPO \cite{zheng2025gspo} reduce training cost by deriving advantages from grouped rollouts.

\paragraph{Joint Optimized Optimization of Reward and Policy Models.}
A fixed reward model exacerbates distribution shifts and reward hacking during online policy optimization \cite{yuan2024self, huang2026real}. Recent work jointly updates the reward and policy: Self-Rewarding LMs \cite{yuan2024self} improve both instruction following and reward modeling via iterative DPO, while R2M \cite{huang2026real} incorporates the policy's hidden states into the reward model for lightweight online adaptation. For math, code, and agentic tasks, PRIME \cite{cui2025process} and iStar \cite{liu2026agentic} alternate between an implicit process reward model and the policy, avoiding costly dense step-level annotations. Overall, the paradigm is shifting from isolated reward modeling toward online joint evolution.


\section{Preliminaries}

Reinforcement learning from human feedback (RLHF) optimizes an autoregressive policy $\pi_\theta(y\mid x)$ to maximize a scalar reward $r(x,y)$ while staying close to a reference policy $\pi_{\mathrm{ref}}$:
\begin{equation}
\begin{aligned}
    \mathcal{J}(\theta) = & \, \mathbb{E}_{x \sim \mathcal{D}_x,\, y \sim \pi_\theta(\cdot \mid x)} \left[ r(x, y) \right] \\
    & - \beta \, \mathbb{KL}\!\left[ \pi_\theta(\cdot \mid x) \,\|\, \pi_{\mathrm{ref}}(\cdot \mid x) \right],
\end{aligned}
\label{eq:rlhf_objective}
\end{equation}
where $\beta>0$ controls KL regularization. 

Since human reward is not directly observed, RLHF uses a learned reward model $r_\xi(x,y)$ trained on pairwise preferences $\mathcal{D}=\{(x,y_w,y_l)\}$ with the Bradley-Terry (BT) objective \cite{bradley1952rank}:
\begin{equation}
\begin{aligned}
    \mathcal{L}_{\text{BT}}(\xi) = - \, \mathbb{E}_{(x, y_w, y_l) \sim \mathcal{D}} \big[ & \log \sigma( r_\xi(x, y_w) \\
    & - r_\xi(x, y_l) ) \big].
\end{aligned}
\label{eq:bt_loss}
\end{equation}
where $y_w$ is preferred over $y_l$ and $\sigma(\cdot)$ is the logistic sigmoid.

For a policy $\pi$, the prompt value is the expected response reward,
\begin{equation}
    V^{\pi}(x) = \mathbb{E}_{y \sim \pi(\cdot \mid x)}\left[ r(x, y) \right].
\end{equation}
A learned value baseline can reduce policy-gradient variance without bias \cite{williams1992simple,sutton2018reinforcement}. Recent LLM RLHF pipelines often avoid training a separate critic by using group-based estimators such as RLOO \cite{ahmadian2024back}, GRPO \cite{shao2024deepseekmath}, and GSPO \cite{zheng2025gspo}. 

\section{On-Policy Feedback for Reward Model Self-Supervised Improvement}
\label{sec:minimax}

In this section, we first introduce a self-supervised reward model training objective by introducing a value-anchored reward model ($\S$\ref{sec:reward_modeling}). Then we reformulate the reinforcement learning objective from maximizing the expected reward to generating informative feedback in $\S\ref{sec:refor_rl}$.

\subsection{Self-Supervised Reward Model Objective}
\label{sec:reward_modeling}
Let $\eta=(\omega,\phi,\psi)$ denote the reward model parameters, where $R_{\omega,\phi}(x,y)$ is the scalar reward and $V_{\omega,\psi}(x)=v_\psi(f_\omega(x))$ is an auxiliary prompt-specific value anchor. $V_{\omega,\psi}$ shares the backbone $f_\omega$, trained to estimate the expected outcome of responses to $x$. This sharing is natural because $f_\omega$ already encodes prompt-level semantics that largely determine the reward distribution, so a lightweight head $v_\psi$ can map these representations to expected reward. We then define the response-level RM advantage as
\begin{equation}
    a^{\mathrm{RM}}_\eta(x,y) = R_{\omega,\phi}(x,y) - V_{\omega,\psi}(x).
\end{equation}
During RL training at step $t$, the policy $\pi_\theta$ samples a response group $Y_i=\{y_{i,1},\dots,y_{i,G}\}$ from $\pi_\theta(\cdot \mid x_i)$ for each prompt $x_i \in X$, and partitions it by the sign of the RM advantage:
\begin{equation}
\begin{aligned}
    C_{\eta,t}(x_i,Y_i)
    &=\{y\in Y_i:a^{\mathrm{RM}}_\eta(x_i,y)\ge 0\},\\
    L_{\eta,t}(x_i,Y_i)
    &=\{y\in Y_i:a^{\mathrm{RM}}_\eta(x_i,y)<0\}.
\end{aligned}
\end{equation}
Prompts whose response groups yield only one non-empty subset (i.e., $C_{\eta,t}=\emptyset$ or $L_{\eta,t}=\emptyset$) are discarded, as they provide no contrastive signal. For each remaining prompt $x_i$ with both $C_{\eta,t}\neq\emptyset$ and $L_{\eta,t}\neq\emptyset$, we train $R_{\omega,\phi}(x,y)$ with the following value-anchored contrastive objective inspired by \citet{wang2025adaptive}:
\begin{equation}
\resizebox{0.95\linewidth}{!}{$
\displaystyle
\begin{aligned}
    \ell_{\mathrm{R}}(\eta;x,Y,t)
    ={}&
    -\frac{1}{|C_{\eta,t}|}
    \sum_{y\in C_{\eta,t}}
    \log
    \frac{
        \exp(R_{\omega,\phi}(x,y))
    }{
        \exp(V_{\omega,\psi}(x))+
        \sum_{y'\in C_{\eta,t}}\exp(R_{\omega,\phi}(x,y'))
    } \\
    &-
    \log
    \frac{
        \exp(V_{\omega,\psi}(x))
    }{
        \exp(V_{\omega,\psi}(x))+
        \sum_{y\in L_{\eta,t}}\exp(R_{\omega,\phi}(x,y))
    } .
\end{aligned}
$}
\label{eq:contrastive_game_loss}
\end{equation}
  And the value anchor $V_{\omega,\psi}(x)$ is optimized on the entire sampled group using the Mean Squared Error (MSE) loss,
\begin{equation}
\resizebox{0.85\linewidth}{!}{$
\displaystyle
    \ell_{\mathrm{V}}(\eta;x,Y)
    =
    \left(
        V_{\omega,\psi}(x)
        -
        \frac{1}{|Y|}
        \sum_{y\in Y}R_{\omega,\phi}(x,y)
    \right)^2 .
$}
\label{eq:value_game_loss}
\end{equation}

\textbf{Remark 1.} For any prompt $x$ and policy $\pi_\theta$, the empirical average $\frac{1}{K}\sum_{j=1}^{K}R_{\omega,\phi}(x,y_j)$ over i.i.d.\ samples $y_j\sim\pi_\theta(\cdot\mid x)$ is an unbiased estimator of the policy-conditioned value
$V^{\pi_\theta}(x)=\mathbb{E}_{y\sim\pi_\theta(\cdot\mid x)}[R_{\omega,\phi}(x,y)]$,
with variance decreasing as $\mathcal{O}(1/K)$. Setting $K=G$ and identifying $\{y_j\}$ with the sampled group $Y$, the MSE target in \cref{eq:value_game_loss} therefore provides an unbiased estimate of $V^{\pi_\theta}(x)$. Minimizing the value-anchor regression objective drives $V_{\omega,\psi}(x)$ toward this policy-conditioned expectation, yielding a calibrated baseline for the subsequent contrastive partition.

Let $\mathcal{P}_{\theta,\eta,t}^{\pm}$ represent the conditional distribution over prompt–response groups for which $C_{\eta,t}(x,Y)\neq\emptyset$ and $L_{\eta,t}(x,Y)\neq\emptyset$. The expected loss of the reward model is then given by
\begin{equation}
\resizebox{0.85\linewidth}{!}{$
\displaystyle
\begin{aligned}
    \mathcal{R}_t(\eta,\theta)
    ={}&
    \mathbb{E}_{(x,Y)\sim\mathcal{P}_{\theta,\eta,t}^{\pm}}
    [\ell_{\mathrm{R}}(\eta;x,Y,t)] \\
    &+
    \lambda_V
    \mathbb{E}_{x\sim\mathcal{D}_x,\;Y\sim\pi_\theta^G(\cdot\mid x)}
    [\ell_{\mathrm{V}}(\eta;x,Y)] ,
\end{aligned}
$}
\label{eq:rm_game_loss}
\end{equation}
where $\lambda_V>0$ controls the value-anchor regression objective.

\subsection{From Maximizing Expected Reward to Generating Informative Feedback}
\label{sec:refor_rl}

In standard RLHF, the policy maximizes the expected reward objective $\mathcal{J}(\theta)$ in \cref{eq:rlhf_objective} while treating the reward model as fixed. We now show that this same policy update, under sufficient local conditions, simultaneously generates informative on-policy feedback for the reward model. To formalize this connection, we combine the expected reward model loss with a KL-regularized policy term to define the joint objective at step $t$:
\begin{equation}
\resizebox{0.85\linewidth}{!}{$
\displaystyle
\begin{aligned}
    \mathcal{M}_t(\eta,\theta)
    ={}&
    \mathcal{R}_t(\eta,\theta) \\
    &-
    \beta_\pi\,
    \mathbb{E}_{x\sim\mathcal{D}_x}
    \left[
        \mathbb{KL}\!\left(\pi_\theta(\cdot\mid x)\,\|\,\pi_{\mathrm{ref}}(\cdot\mid x)\right)
    \right].
\end{aligned}
$}
\label{eq:minimax_objective}
\end{equation}
The following lemma shows that the standard policy gradient direction for $\mathcal{J}(\theta)$ is also a local ascent direction for \cref{eq:minimax_objective} under the reward over-optimization regime. We defer the full assumptions to \cref{sec:proof_lemma1}. Informally, policy optimization should expose RM epistemic errors, the RM loss should be locally sensitive to these errors, and the KL term should not dominate this effect.

\textbf{Lemma 1}. \textit{Fix $\eta$ and $t$, and let $g_\theta=\nabla_\theta\mathcal{J}(\theta)$. Under Conditions~(C1)--(C3) in \cref{sec:proof_lemma1}, $g_\theta$ is a local ascent direction for $\mathcal{M}_t(\eta,\theta)$ with respect to the policy parameters:
$\langle \nabla_\theta \mathcal{J}(\theta),\, \nabla_\theta \mathcal{M}_t(\eta,\theta) \rangle > 0$.}
(Proof in \cref{sec:proof_lemma1})

Thus, reward maximization naturally produces hard on-policy examples for the current reward model.

This yields the reward-model-centric minimax objective:
\begin{equation}
    \min_{\eta}\; \max_{\theta}\; \mathcal{M}_t(\eta,\theta).
    \label{eq:minimax_game}
\end{equation}

In this minimax formulation, the policy seeks KL-constrained responses that expose weaknesses of the current reward model, while the reward model is updated to reduce ranking and calibration errors on these samples.

\textbf{Proposition 1.} \textit{Under the conditions of Lemma~1, value-anchored on-policy feedback stochastically approximates the minimax problem in \cref{eq:minimax_game}: the reward model step descends a stop-gradient surrogate of the outer objective, while the policy step approximately ascends the inner objective.} (Proof in \cref{sec:proof_prop1})

\begin{algorithm*}[!t]
\caption{Self-Supervised Reward Model Improvement via Value-Anchored On-Policy Feedback}
\label{alg:co_training}
\textbf{Input} Instruction set $X$, initial policy $\pi_{\theta_0}$, initial reward model $(\omega_0,\phi_0,\psi_0)$, initial margin $m_0$, training steps $T$, group size $G$.
\begin{algorithmic}[1]
\For{training step $t = 1$ \textbf{to} $T$}
    \State Set curriculum threshold $\mu(t)$ from $m_0$ and sample a batch $B \subset X$
    
    \Comment{\textit{Step 1: Adaptive Feedback Filtering}}
    
    \For{each instruction $x_i \in B$}
        \State Sample $G$ responses $Y_i \sim \pi_{\theta_{t-1}}(\cdot \mid x_i)$
        \State Compute RM advantages $a^{\mathrm{RM}}_{i,j}=R_{\omega_{t-1},\phi_{t-1}}(x_i,y_{i,j})-V_{\omega_{t-1},\psi_{t-1}}(x_i)$
        \State Filter informative responses with $\sigma(|a^{\mathrm{RM}}_{i,j}|)\ge\mu(t)$ and split them into $\tilde{Y}_i^{+}$ and $\tilde{Y}_i^{-}$
    \EndFor
    \State Keep valid prompts $\tilde{B}=\{x_i\in B\mid \tilde{Y}_i^{+}\neq\emptyset \wedge \tilde{Y}_i^{-}\neq\emptyset\}$
    
    \Comment{\textit{Step 2: Self-Supervised Reward Model Improvement}}
    
    \State Update $(\omega_t,\phi_t)$ with reward loss $\mathcal{L}_R$ and update $\psi_t$ with value loss $\mathcal{L}_V$
    
    \Comment{\textit{Step 3: Policy Model Optimization}}
    
    \State Recompute RM advantages, filtering, and rewards on the same sampled groups using $(\omega_t,\phi_t,\psi_t)$, then optimize $\pi_{\theta_{t-1}}\to\pi_{\theta_t}$ with a standard RL algorithm
\EndFor
\end{algorithmic}
\textbf{Output} Improved reward model $(\omega_T,\phi_T,\psi_T)$ and optimized policy $\pi_{\theta_T}$
\end{algorithm*}

\section{Methodology}

This section instantiates the minimax formulation in $\S$\ref{sec:minimax} as \method. We first initialize the value-anchored reward model ($\S$\ref{subsec:rm_init}) and then describe the on-policy feedback loop ($\S$\ref{sec:co-training}). \cref{fig:savor_overview,alg:co_training} give the overview and full procedure.

\subsection{Value-Anchored Reward Modeling} \label{subsec:rm_init}

We instantiate the value-anchored reward model in \cref{sec:reward_modeling} with a shared backbone $f_\omega$, a reward head $h_\phi$, and a value head $v_\psi$. Initialization has two stages: preference learning for $(\omega,\phi)$ and value anchor integration for $\psi$.

\textbf{Stage 1: Preference Learning.} We jointly train $f_\omega$ and $h_\phi$ on pairwise preferences $\mathcal{D}=\{(x,y_w,y_l)\}$ using the standard Bradley-Terry (BT) objective \cite{bradley1952rank}, with $R_{\omega,\phi}(x,y)=h_\phi(f_\omega(x,y))$:
\begin{equation}
\resizebox{0.85\linewidth}{!}{$
    \displaystyle
    \mathcal{L}_{\text{BT}}(\omega, \phi) = -\log \sigma(R_{\omega, \phi}(x, y_w) - R_{\omega, \phi}(x, y_l)).
$}
\end{equation}

\textbf{Stage 2: Value Anchor Integration.} We then freeze $(\omega,\phi)$ and fit the value head on prompt-only data $\mathcal{D}_x=\{x_i\}$. For each prompt $x$, a sampling policy $\pi_{\theta_{\mathrm{s}}}$ draws $K$ responses $y_j\sim\pi_{\theta_{\mathrm{s}}}(\cdot\mid x)$, the mean frozen-RM score is used as the target for $V_{\omega,\psi}(x)=v_\psi(f_\omega(x))$:
\begin{equation}
\resizebox{0.8\linewidth}{!}{$
\displaystyle
    \mathcal{L}_{V}(\psi) =
    \left(
        V_{\omega, \psi}(x) -
        \frac{1}{K}\sum_{j=1}^{K} R_{\omega, \phi}(x, y_j)
    \right)^2.
$}
\label{eq:value_learning}
\end{equation}

Thus \cref{eq:value_learning} is the finite-sample initialization counterpart of the value-regression term in \cref{eq:value_game_loss}. By Remark~1, its target is an unbiased estimator of the policy-conditioned value under $\pi_{\theta_{\mathrm{s}}}$, providing the anchor reused during on-policy feedback in \cref{sec:co-training}.

\subsection{Value-Anchored On-Policy Feedback} \label{sec:co-training}

\method improves the reward model from the policy's own samples through three stages per batch: adaptive feedback filtering, self-supervised reward model improvement, and policy optimization with the improved rewards. At training step $t$, we sample a batch of instructions $B \subset X$ and proceed as follows.

\textbf{Step 1: Adaptive Feedback Filtering.} For each instruction $x_i \in B$, we draw a group of $G$ candidate responses $Y_i = \{y_{i, 1}, \dots, y_{i, G}\}$ from the current policy $\pi_{\theta_{t-1}}$. For each response, we compute its response-level RM advantage under the current reward model,
\begin{equation}
\resizebox{0.85\linewidth}{!}{$
    \displaystyle
    a^{\mathrm{RM}}_{t-1}(x_i,y_{i,j})
    =
    R_{\omega_{t-1}, \phi_{t-1}}(x_i, y_{i,j})
    -
    V_{\omega_{t-1}, \psi_{t-1}}(x_i).
$}
\end{equation}

Following curriculum learning \cite{bengio2009curriculum}, we remove near-zero-advantage responses with a dynamic threshold $\mu(t)$:
\begin{equation}
\resizebox{0.85\linewidth}{!}{$
    \displaystyle
    p(t) = \min\left(1 - m_0 + \frac{t}{T}, 1\right), \quad \mu(t) = \frac{2 - p(t)}{2},
$}
\end{equation}
where $T$ is the total number of training steps and $m_0$ is the initial margin. A sampled response $y_{i,j}$ is retained in $\tilde{Y}_i$ iff
\begin{equation}
    s = \mathbb{I} \left[ \sigma \left( \left| a^{\mathrm{RM}}_{t-1}(x_i,y_{i,j}) \right| \right) \ge \mu(t) \right].
\label{eq:response_filter}
\end{equation}
Since $\mu(t)$ decays over training, early updates emphasize clearly separated responses while later updates admit broader on-policy feedback.

We partition the retained responses by the sign of their RM advantage:
\begin{equation}
    \begin{aligned}
        \tilde{Y}_i^{+} &= \{ y \in \tilde{Y}_i \mid a^{\mathrm{RM}}_{t-1}(x_i,y) \geq 0 \}, \\
        \tilde{Y}_i^{-} &= \{ y \in \tilde{Y}_i \mid a^{\mathrm{RM}}_{t-1}(x_i,y) < 0 \}.
    \end{aligned}
\end{equation}
Only prompts with both positive and negative subsets are kept:
\begin{equation}
    \tilde{B} =
    \left\{
        x_i \in B
        \mid
        \tilde{Y}_i^{+} \neq \emptyset
        \;\wedge\;
        \tilde{Y}_i^{-} \neq \emptyset
    \right\}.
\end{equation}

\textbf{Step 2: Self-Supervised Reward Model Improvement.} Given $\tilde{B}$, we initialize $(\omega_t,\phi_t,\psi_t)\gets(\omega_{t-1},\phi_{t-1},\psi_{t-1})$ and optimize a stop-gradient implementation of the reward model objective in \cref{eq:contrastive_game_loss,eq:value_game_loss}. The reward loss updates $(\omega,\phi)$ while treating the value anchor as fixed:
\begin{equation}
    \mathcal{L}_R(\omega_t, \phi_t) = \frac{1}{|\tilde{B}|} \sum_{x_i \in \tilde{B}} \left( \mathcal{L}_{+, i} + \mathcal{L}_{-, i} \right),
\label{eq:continual_reward_loss}
\end{equation}
where the positive and negative RM advantage terms take the softmax forms
\begin{equation}
\resizebox{0.85\linewidth}{!}{$
\displaystyle
    \mathcal{L}_{+, i} = -\frac{1}{|\tilde{Y}_i^{+}|}\!\!\sum_{y \in \tilde{Y}_i^{+}}\!\! \log \frac{\exp\!\left(R_{\omega_t, \phi_t}(x_i, y)\right)}{\exp\!\left(\mathbf{SG}[V_{\omega_t, \psi_t}(x_i)]\right) + \sum_{y' \in \tilde{Y}_i^{+}} \exp\!\left(R_{\omega_t, \phi_t}(x_i, y')\right)},
$}
\label{eq:positive_reward_loss}
\end{equation}
\begin{equation}
\resizebox{0.85\linewidth}{!}{$
\displaystyle
    \mathcal{L}_{-, i} = -\log \frac{\exp\!\left(\mathbf{SG}[V_{\omega_t, \psi_t}(x_i)]\right)}{\exp\!\left(\mathbf{SG}[V_{\omega_t, \psi_t}(x_i)]\right) + \sum_{y \in \tilde{Y}_i^{-}} \exp\!\left(R_{\omega_t, \phi_t}(x_i, y)\right)}.
$}
\label{eq:negative_reward_loss}
\end{equation}
These terms move positive advantage responses above the anchor and negative advantage responses below it. The value head is then calibrated to the full sampled group with:
\begin{equation}
\resizebox{0.85\linewidth}{!}{$
\displaystyle
    \mathcal{L}_V(\psi_t) = \frac{1}{|B|} \sum_{x_i \in B} \left( V_{\mathbf{SG}[\omega_t], \psi_t}(x_i) - \frac{1}{|Y_i|} \sum_{y \in Y_i} \mathbf{SG}[R_{\omega_t, \phi_t}(x_i, y)] \right)^2.
$}
\label{eq:continual_value_loss}
\end{equation}
The stop-gradient operators separate the two updates: $\mathcal{L}_R$ affects only $(\omega,\phi)$, while $\mathcal{L}_V$ affects only $\psi$, yielding the improved reward model $(\omega_t,\phi_t,\psi_t)$.

\textbf{Step 3: Policy Model Optimization.} We reuse the sampled response groups $\{Y_i\}_{x_i\in B}$ from Step 1 and recompute RM advantages, filtering, and rewards with the updated $(\omega_t,\phi_t,\psi_t)$, obtaining $\{\tilde{Y}_{i,\text{new}}\}_{x_i\in B}$ and $R_{\text{new}}$. The policy is then updated from $\pi_{\theta_{t-1}}$ to $\pi_{\theta_t}$ via a standard RL algorithm on these filtered responses, producing the next on-policy distribution.

\begin{table*}[t]
    \centering
    \renewcommand{\arraystretch}{1.12}
    \resizebox{\linewidth}{!}{
    \begin{tabular}{l c c c c c c c}
        \toprule
        \textbf{Method} & \textbf{RewardBench} & \textbf{RewardBench 2} & \textbf{RM-Bench} & \textbf{\makecell{PPE \\ Preference}} & \textbf{\makecell{PPE \\ Correctness}} & \textbf{JudgeBench} & \textbf{Average} \\
        \midrule
        Skywork-Reward-V2-Llama-3.2-3B & 93.0 & 74.7 & 80.0 & 68.4 & 70.7 & 69.2 & 76.0 \\
        Continual Offline Training RM & 93.1 & 75.3 & 80.6 & 68.5 & 71.0 & 70.0 & 76.4 \\
        \midrule
        \multicolumn{8}{c}{\textit{Policy Model: Qwen2.5-3B-Instruct}} \\
        HL-BT & \underline{93.2} & 74.9 & 81.0 & \textbf{68.4} & 70.5 & 67.7 & 76.0 \\
        Mean Reward & 83.1 & 65.0 & 76.6 & 62.4 & 65.9 & 60.9 & 69.0 \\
        \method & \textbf{93.6} & \textbf{76.0} & \textbf{82.1} & 67.5 & \textbf{71.1} & \textbf{70.0} & \textbf{76.7} \\
        \quad w/o Curriculum Mechanism & \textbf{93.6} & \underline{75.9} & \underline{81.9} & 67.5 & \underline{71.0} & \underline{69.1} & \underline{76.5} \\
        \quad w/o Policy Model Optimization & \textbf{93.6} & 75.7 & 81.7 & \underline{67.9} & 70.9 & 68.3 & 76.4 \\
        \midrule
        \multicolumn{8}{c}{\textit{Policy Model: Qwen3-4B-Instruct-2507}} \\
        HL-BT & 93.3 & 75.1 & 81.4 & 68.3 & 70.5 & 67.7 & 76.1 \\
        Mean Reward & 83.4 & 65.0 & 75.8 & 60.0 & 66.6 & 60.9 & 68.6 \\
        \method & \textbf{93.9} & \textbf{76.1} & \textbf{82.3} & \textbf{68.6} & \textbf{71.2} & \textbf{71.4} & \textbf{77.3} \\
        \quad w/o Curriculum Mechanism & \underline{93.6} & \textbf{76.1} & \underline{82.0} & \underline{67.8} & \textbf{71.2} & \underline{68.9} & \underline{76.6} \\
        \quad w/o Policy Model Optimization & 93.5 & \underline{76.0} & \underline{82.0} & \underline{67.8} & \underline{71.1} & 68.3 & 76.5 \\
        \bottomrule
    \end{tabular}}
    \caption{Reward model evaluation across six benchmarks. All on-policy methods use GRPO for response generation. The best results per policy model are in \textbf{bold}, the second best are \underline{underlined}.}
    \label{tab:benchmark_results}
\end{table*}

\begin{table*}[t]
    \centering
    \renewcommand{\arraystretch}{1.12}
    \resizebox{\linewidth}{!}{
    \begin{tabular}{l c c c c c c c}
        \toprule
        \textbf{RL Algorithm} & \textbf{Reward Bench} & \textbf{Reward Bench 2} & \textbf{RM-Bench} & \textbf{\makecell{PPE \\ Preference}} & \textbf{\makecell{PPE \\ Correctness}} & \textbf{JudgeBench} & \textbf{Average} \\
        \midrule
        \multicolumn{8}{c}{\textit{Policy Model: Qwen2.5-3B-Instruct}} \\
        \method with GRPO & \textbf{93.6} & \textbf{76.0} & \textbf{82.1} & 67.5 & \textbf{71.1} & \textbf{70.0} & \textbf{76.7} \\
        \method with RLOO & 93.5 & 75.9 & 81.9 & 67.5 & 71.0 & 68.6 & 76.4 \\
        \method with GSPO & 93.5 & \textbf{76.0} & \textbf{82.1} & \textbf{67.6} & 71.0 & 69.1 & 76.6 \\
        \midrule
        \multicolumn{8}{c}{\textit{Policy Model: Qwen3-4B-Instruct-2507}} \\
        \method with GRPO & \textbf{93.9} & 76.1 & 82.3 & 68.6 & \textbf{71.2} & \textbf{71.4} & \textbf{77.3} \\
        \method with RLOO & 93.5 & \textbf{76.2} & 81.9 & \textbf{68.7} & 71.1 & 69.4 & 76.8 \\
        \method with GSPO & 93.8 & 75.4 & \textbf{82.4} & 68.2 & 71.1 & 68.3 & 76.5 \\
        \bottomrule
    \end{tabular}}
    \caption{Effect of the RL algorithm choice on reward model improvement under \method.}
    \label{tab:rl_algorithm_comparison}
\end{table*}

\section{Experiments}

\subsection{Setup}
\paragraph{Models and RL Algorithms.}
We use Skywork-Reward-V2-Llama-3.2-3B \cite{SkyworkRewardV2} as the initial reward model backbone, which is trained on pairwise preference data with the Bradley-Terry objective on Llama-3.2-3B-Instruct \cite{Llama3}. We augment it with a value head as described in \cref{subsec:rm_init}. For the policy model, we experiment with two instruction-tuned backbones: Qwen2.5-3B-Instruct \cite{qwen2.5} and Qwen3-4B-Instruct-2507 \cite{qwen3technicalreport}. For policy optimization, we consider three RL algorithms: GRPO \cite{shao2024deepseekmath}, RLOO \cite{ahmadian2024back}, and GSPO \cite{zheng2025gspo}.

\paragraph{Data.}
We use UltraFeedback \cite{cui2024ultrafeedback} as the prompt source for value integration and policy optimization. During RL training, only prompts are used; responses and reward scores are generated online by the policy and RM as the on-policy feedback in \cref{sec:co-training}. More details can be seen in  \cref{sec:more_training_details}.

\paragraph{Reward Model Evaluation.}
We evaluate the reward model on six established benchmarks: RewardBench \cite{rewardbench} and RewardBench 2 \cite{rewardbench2} for preference-based ranking accuracy, RM-Bench \cite{RM-Bench} for robustness to stylistic bias and subtle quality differences, PPE Preference and PPE Correctness \cite{PPE} for preference alignment and correctness assessment, and JudgeBench \cite{JudgeBench} for judging complex real-world responses.

\paragraph{Policy Model Evaluation.}
To evaluate whether improved reward models translate into better downstream policy performance, we assess the trained policy models on two open-ended instruction-following benchmarks: AlpacaEval 2 \cite{alpaca_eval}, which reports length-controlled win rate (LC) \cite{dubois2024length} and raw win rate (WR) against GPT-5.2, and Arena-Hard-v2.0 \cite{li2024crowdsourced}, which evaluates performance on challenging real-world instructions. Both benchmarks use GPT-4.1-mini as the judge.\footnote{We use \texttt{gpt-5.2-2025-12-11} as the AlpacaEval 2 reference model and \texttt{gpt-4.1-mini-2025-04-14} as the judge for both benchmarks.}

\paragraph{Reward Model Baselines.}
We compare reward modeling of \method against the following baselines: (1) the initial reward model without any online updating; (2) continual offline training RM, which continues training the reward model on static preference data\footnote{We use \texttt{HuggingFaceH4/ultrafeedback\_binarized} for continual offline training.}; (3) Mean Reward, which directly uses the mean reward of the on-policy response group as the value estimate without learned value decomposition; and (4) HL-BT (Highest and Lowest BT Model), which directly pairs the highest and lowest reward responses from each on-policy group and updates the reward model with the Bradley-Terry loss in \cref{eq:bt_loss}.

\paragraph{Policy Optimization Baselines.}
For policy optimization, we additionally compare against: (1) PRIME \cite{cui2025process}, which alternates between training an implicit process reward model and the policy via online reinforcement learning; and (2) R2M \cite{huang2026real}, which improves the performance of the reward model through real-time hidden states from policy model.

\subsection{Main Results}

\begin{table}[t]
\centering
\renewcommand{\arraystretch}{1.12}
\resizebox{\linewidth}{!}{
\begin{tabular}{lccc}
\toprule
\multirow{2}{*}{\textbf{Method}} & \multicolumn{2}{c}{\textbf{AlpacaEval 2}} & \textbf{Arena-Hard-v2.0} \\
\cmidrule(lr){2-3} \cmidrule(lr){4-4}
& LC (\%) & WR (\%) & Win Rate (\%) \\
\midrule
\multicolumn{4}{c}{\textit{Policy Model: Qwen2.5-3B-Instruct}} \\
SFT    & 15.36 & 18.41 & 2.0 \\
REINFORCE++   &   15.98    &  19.10     &  2.1   \\
PRIME & 12.18 & 14.33 & 1.3 \\
\cmidrule(lr){1-4}
GRPO          & 16.31 & \underline{24.41} & \underline{2.2} \\
\quad + R2M    & 17.45 & 21.05 & 1.9 \\
\quad + \method & \underline{17.85} & 21.59 & \underline{2.2} \\
\quad + Improved RM & \textbf{20.20} & \textbf{27.72} & \textbf{2.6} \\
\midrule
\multicolumn{4}{c}{\textit{Policy Model: Qwen3-4B-Instruct-2507}} \\
SFT    & 50.01 & 58.59 & 31.9 \\
REINFORCE++   &   49.04    &   55.81    &   28.6   \\
PRIME & 43.37 & 52.83 & 22.7 \\
\cmidrule(lr){1-4}
GRPO          & 51.68 & 59.81 & 30.2 \\
\quad + R2M    & 52.00 & 62.26 & 32.6 \\
\quad + \method & \underline{53.28} & \textbf{62.69} & \underline{33.5} \\
\quad + Improved RM & \textbf{54.24} & \underline{62.40} & \textbf{33.9} \\
\bottomrule
\end{tabular}}
\caption{Downstream policy performance on AlpacaEval 2 and Arena-Hard-v2.0. The best results per policy model are in \textbf{bold}, the second best are \underline{underlined}.}
\label{tab:main_results}
\end{table}

\paragraph{Reward Model Evaluation.}
\cref{tab:benchmark_results} summarizes reward model performance across six benchmarks. Starting from Skywork-Reward-V2-Llama-3.2-3B, \method consistently improves RM capability regardless of the policy used to generate on-policy feedback. With Qwen2.5-3B-Instruct, \method increases the average score from 76.0 to 76.7 and achieves the best results on most benchmarks. Using the stronger Qwen3-4B-Instruct-2507 further improves the average to 77.3 and obtains the best scores on all six benchmarks, suggesting that stronger policies provide more informative on-policy samples for RM self-improvement.

Continual offline training on static preference data yields only marginal gains over the initial RM, confirming that the reward model benefits more from on-policy feedback aligned with the evolving policy distribution than from additional passes over fixed preference data. The Mean Reward baseline suffers severe degradation, indicating that a learned value decomposition is essential for deriving reliable self-supervised signals. The HL-BT baseline matches the initial RM on average but consistently underperforms \method. This is because even when the highest and lowest reward responses are semantically equivalent (e.g., both correct or incorrect), HL-BT still forces the reward model to enlarge the gap between them, causing it to overfit to superficial structural differences rather than genuine semantic quality and thus increasing susceptibility to reward hacking (see \cref{tab:case_study} for a concrete example). This gap highlights the advantage of value-anchored contrastive learning with adaptive filtering over naive pairwise ranking on extreme responses.

\paragraph{Ablation Study.}
The ablation results in \cref{tab:benchmark_results} reveal the contribution of each component. Removing the curriculum mechanism retains competitive scores on RewardBench and RewardBench 2 but degrades average scores, dropping by 0.7 and 0.2 on Qwen3-4B-Instruct-2507 and Qwen2.5-3B-Instruct, respectively. This indicates that curriculum-driven filtering is critical for complex evaluation scenarios where noisy or ambiguous self-supervised signals would otherwise mislead the reward model. Freezing the policy further reduces performance: the average drops from 77.3 to 76.5 with Qwen3-4B-Instruct-2507 and from 76.7 to 76.4 with Qwen2.5-3B-Instruct. This corroborates Lemma~1, which shows that policy optimization naturally steers on-policy responses toward regions where the RM is most miscalibrated, freezing the policy disables this adaptive feedback generation and limits the RM's self-improvement.

\paragraph{Effect of RL Algorithms.}
As shown in \cref{tab:rl_algorithm_comparison}, \method consistently improves reward model performance across GRPO, RLOO, and GSPO under both policy backbones. Because all three algorithms produce grouped on-policy responses that can be transformed into value-anchored feedback, these gains transfer across policy update algorithms, indicating that \method is largely optimizer-agnostic.

\paragraph{Policy Model Evaluation.}
\cref{tab:main_results} reports downstream policy performance. We evaluate \method under co-training (``+ \method'') and by using the improved reward model to train a fresh policy (``+ Improved RM''). Both variants consistently outperform vanilla GRPO across benchmarks and policy backbones, showing that better reward modeling translates to stronger policies. The improved RM setting achieves the best overall results, indicating that the learned reward model is transferable beyond the co-training process. Compared with R2M, \method remains competitive or stronger without coupling the reward model to policy hidden states. PRIME underperforms because it relies on verifiable rewards to train an implicit process reward model, while open-ended instruction-following tasks lack absolute correct or incorrect rewards, making such training less effective.

\section{Conclusion}

We present \method, a self-supervised framework that uses on-policy responses to continuously improve reward models without extra human labels or external judges. With a prompt-specific value anchor, adaptive feedback filtering, and value-anchored contrastive learning, \method turns policy rollouts into reliable RM feedback. Experiments show consistent RM gains across benchmarks and stronger downstream policy performance, suggesting that on-policy feedback is an effective supervision for improving reward models and policies.

\clearpage

\section*{Limitations}

Although \method shows consistent improvements across policy backbones and critic-free RL algorithms, several limitations remain. First, our experiments use policy models at the 3B to 4B scale and a 3B reward model backbone, so further work is needed to verify whether the same training dynamics hold for substantially larger models. Second, our evaluation mainly relies on automatic reward model benchmarks and LLM-based judges. Human preference studies would provide stronger evidence, especially for safety-critical, culturally sensitive, or highly subjective instructions. Finally, \method introduces additional computation because it samples multiple responses per prompt and updates the reward model during RL training. Improving the efficiency of this feedback loop is an important direction for future work.



\bibliography{custom}

\clearpage

\appendix

\section{Proof of Lemma 1}
\label{sec:proof_lemma1}

\paragraph{Formal assumptions.}
We state the assumptions used in Lemma~1. Let $R_{\omega,\phi}(x,y)=R^*(x,y)+\epsilon(x,y)$, where $R^*$ is the latent true reward and $\epsilon$ is the epistemic error. Fix $\eta$ and $t$, and let $g_\theta=\nabla_\theta\mathcal{J}(\theta)$. We assume this local argument is taken at a point where $S$, $K$, and $\mathcal{R}_t$ are differentiable with respect to $\theta$; equivalently, the point is away from sign-partition boundaries, or a standard smoothing/dominated-convergence argument justifies the directional derivatives below. Define the on-policy epistemic-error dispersion
\begin{equation}
    S(\theta)=
    \mathbb{E}_{x\sim\mathcal{D}_x,\,y\sim\pi_\theta(\cdot\mid x)}
    [\epsilon(x,y)^2],
\end{equation}
and the KL term
\begin{equation}
    K(\theta)=
    \mathbb{E}_{x\sim\mathcal{D}_x}
    \left[
        \mathbb{KL}\!\left(\pi_\theta(\cdot\mid x)\,\|\,\pi_{\mathrm{ref}}(\cdot\mid x)\right)
    \right].
\end{equation}
For any locally differentiable function $F$, write
\begin{equation}
    D_{g_\theta}^{+}F(\theta)
    =
    \frac{d}{d\alpha}F(\theta+\alpha g_\theta)\Big|_{\alpha=0^+}.
\end{equation}
Assume that:
\begin{itemize}
    \item[(C1)] \textbf{Over-optimization bias.} Moving along $g_\theta$ increases the expected squared epistemic error under the on-policy distribution:
    \begin{equation}
        D_{g_\theta}^{+}S(\theta)>0.
    \end{equation}
    \item[(C2)] \textbf{Loss sensitivity.} The reward model risk in \cref{eq:rm_game_loss} is locally sensitive to this dispersion: there exists $c_t>0$ such that
    \begin{equation}
        D_{g_\theta}^{+}\mathcal{R}_t(\eta,\theta)
        \ge c_t\,D_{g_\theta}^{+}S(\theta).
    \end{equation}
    \item[(C3)] \textbf{Local regularization balance.} The KL change does not offset this increase:
    \begin{equation}
        \beta_\pi D_{g_\theta}^{+}K(\theta)
        < c_t\,D_{g_\theta}^{+}S(\theta).
    \end{equation}
\end{itemize}

\textit{Proof.}
It suffices to prove $D_{g_\theta}^{+}\mathcal{M}_t(\eta,\theta)>0$, because by the chain rule
\begin{equation}
    D_{g_\theta}^{+}\mathcal{M}_t(\eta,\theta)
    =
    \langle g_\theta,\nabla_\theta\mathcal{M}_t(\eta,\theta)\rangle .
\end{equation}

By \cref{eq:minimax_objective}, the inner objective can be written as
\begin{equation}
    \mathcal{M}_t(\eta,\theta)
    =
    \mathcal{R}_t(\eta,\theta)-\beta_\pi K(\theta),
\end{equation}
Taking the directional derivative along $g_\theta$ gives
\begin{equation}
    D_{g_\theta}^{+}\mathcal{M}_t(\eta,\theta)
    =
    D_{g_\theta}^{+}\mathcal{R}_t(\eta,\theta)
    -
    \beta_\pi D_{g_\theta}^{+}K(\theta).
\end{equation}
Condition~(C1) gives $D_{g_\theta}^{+}S(\theta)>0$. By Condition~(C2),
\begin{equation}
    D_{g_\theta}^{+}\mathcal{R}_t(\eta,\theta)
    \ge
    c_tD_{g_\theta}^{+}S(\theta).
\end{equation}
Combining this inequality with Condition~(C3) yields
\begin{equation}
\begin{aligned}
    D_{g_\theta}^{+}\mathcal{M}_t(\eta,\theta)
    &=
    D_{g_\theta}^{+}\mathcal{R}_t(\eta,\theta)
    -
    \beta_\pi D_{g_\theta}^{+}K(\theta) \\
    &\ge
    c_tD_{g_\theta}^{+}S(\theta)
    -
    \beta_\pi D_{g_\theta}^{+}K(\theta)
    > 0 .
\end{aligned}
\end{equation}
Therefore
\begin{equation}
    \langle \nabla_\theta\mathcal{J}(\theta),
    \nabla_\theta\mathcal{M}_t(\eta,\theta)\rangle
    =
    D_{g_\theta}^{+}\mathcal{M}_t(\eta,\theta)
    >0,
\end{equation}
so $g_\theta$ is a local ascent direction for the inner objective.

\begin{table*}[t]
    \centering
    \renewcommand{\arraystretch}{1.3}
    \resizebox{\linewidth}{!}{
    \begin{tabular}{cc c c c c c c c}
        \toprule
        \textbf{\makecell{Backbone $f_\omega$ \& Reward \\ Head $h_\phi$ LR}} & \textbf{\makecell{Value \\ Head $v_\psi$ LR}} & \textbf{RewardBench} & \textbf{RewardBench 2} & \textbf{RM-Bench} & \textbf{\makecell{PPE \\ Preference}} & \textbf{\makecell{PPE \\ Correctness}} & \textbf{JudgeBench} & \textbf{Average} \\
        \midrule
        \multirow{2}{*}{$1 \times 10^{-6}$} & $1 \times 10^{-5}$ & \textbf{93.7} & 75.6 & \textbf{82.1} & 67.3 & 70.5 & 68.9 & 76.4 \\
         & $1 \times 10^{-6}$ & 93.5 & 75.9 & 81.9 & 67.6 & 71.0 & 68.6 & 76.4 \\
        \midrule
        \multirow{3}{*}{$1 \times 10^{-7}$} & $1 \times 10^{-5}$ & 93.6 & \textbf{76.0} & \textbf{82.1} & 67.5 & \textbf{71.1} & \textbf{70.0} & \textbf{76.7} \\
         & $1 \times 10^{-6}$ & 93.6 & \textbf{76.0} & 81.8 & 67.8 & \textbf{71.1} & 68.3 & 76.4 \\
         & $1 \times 10^{-7}$ & 93.5 & \textbf{76.0} & 81.7 & \textbf{68.1} & \textbf{71.1} & 68.6 & 76.5 \\
        \bottomrule
    \end{tabular}}
    \caption{Sensitivity to learning rates of the backbone \& reward head ($f_\omega$, $h_\phi$) and the value head ($v_\psi$) under \method. All experiments use Qwen2.5-3B-Instruct as the policy model with GRPO.}
    \label{tab:lr_ablation}
\end{table*}

\begin{table*}[t]
    \centering
    \renewcommand{\arraystretch}{1.2}
    \resizebox{\linewidth}{!}{
    \begin{tabular}{c c c c c c c c}
        \toprule
        \textbf{$G$} & \textbf{Reward Bench} & \textbf{Reward Bench 2} & \textbf{RM-Bench} & \textbf{\makecell{PPE \\ Preference}} & \textbf{\makecell{PPE \\ Correctness}} & \textbf{JudgeBench} & \textbf{Average} \\
        \midrule
        4 & \textbf{93.7} & 75.9 & 81.8 & 67.5 & 70.8 & 68.9 & 76.4 \\
        8 & 93.6 & \textbf{76.2} & 81.9 & \textbf{67.7} & 71.0 & 68.9 & 76.6 \\
        16 & 93.6 & 76.0 & \textbf{82.1} & 67.5 & \textbf{71.1} & \textbf{70.0} & \textbf{76.7} \\
        \bottomrule
    \end{tabular}}
    \caption{Effect of group size $G$ (number of responses sampled per prompt) on reward model improvement under \method. All experiments use Qwen2.5-3B-Instruct as the policy model with GRPO.}
    \label{tab:group_size_ablation}
\end{table*}

\begin{table*}[t]
    \centering
    \renewcommand{\arraystretch}{1.3}
    \resizebox{\linewidth}{!}{
    \begin{tabular}{c c c c c c c c}
        \toprule
        \textbf{$m_0$} & \textbf{RewardBench} & \textbf{RewardBench 2} & \textbf{RM-Bench} & \textbf{\makecell{PPE \\ Preference}} & \textbf{\makecell{PPE \\ Correctness}} & \textbf{JudgeBench} & \textbf{Average} \\
        \midrule
        0.3 & \textbf{93.7} & 75.4 & \textbf{82.5} & 66.8 & 70.6 & 69.1 & 76.4 \\
        0.5 & 93.6 & \textbf{76.0} & 82.1 & \textbf{67.5} & \textbf{71.1} & \textbf{70.0} & \textbf{76.7} \\
        0.7 & \textbf{93.7} & \textbf{76.0} & 82.1 & 67.3 & 70.8 & 69.1 & 76.5 \\
        \bottomrule
    \end{tabular}}
    \caption{Sensitivity to the initial margin $m_0$ for curriculum-based filtering under \method. All experiments use Qwen2.5-3B-Instruct as the policy model with GRPO.}
    \label{tab:margin_ablation}
\end{table*}

\section{Proof of Proposition 1}
\label{sec:proof_prop1}

\textit{Proof.}
For fixed $(\eta,\theta,t)$, the sign partition is deterministic after sampling. Thus, the only randomness in a minibatch estimate comes from sampling prompts and response groups. Given a minibatch $B$, sampled groups $\{Y_i\}_{x_i\in B}$, and the subset $\tilde{B}$ of prompts whose sampled groups contain both positive- and negative-RM-advantage responses, define the empirical risk for minibatches with $\tilde{B}\neq\emptyset$ as
\begin{equation}
\begin{aligned}
    \widehat{\mathcal{R}}_t(\eta,\theta)
    ={}&
    \frac{1}{|\tilde{B}|}\sum_{x_i\in\tilde{B}}
    \ell_{\mathrm{R}}(\eta;x_i,Y_i,t) \\
    &+
    \lambda_V
    \frac{1}{|B|}\sum_{x_i\in B}
    \ell_{\mathrm{V}}(\eta;x_i,Y_i).
\end{aligned}
\end{equation}
Conditioned on the event $\tilde{B}\neq\emptyset$, the contrastive term is a Monte Carlo estimate of the conditional risk in \cref{eq:rm_game_loss}: the contrastive average uses groups with both sides of the anchor, while the value term is estimated on all sampled groups. If $\tilde{B}=\emptyset$, the contrastive sub-step is skipped and only the value term is used. The corresponding empirical minimax objective is
\begin{equation}
\begin{aligned}
    \widehat{\mathcal{M}}_t(\eta,\theta)
    ={}&
    \widehat{\mathcal{R}}_t(\eta,\theta)
    -
    \beta_\pi
    \widehat{\mathbb{KL}}(\pi_\theta,\pi_{\mathrm{ref}}).
\end{aligned}
\end{equation}
Let
\begin{equation}
    \widehat{\mathcal{R}}^{\mathrm{ctr}}_t(\eta,\theta)
    =
    \frac{1}{|\tilde{B}|}\sum_{x_i\in\tilde{B}}
    \ell_{\mathrm{R}}(\eta;x_i,Y_i,t).
\end{equation}
By linearity of expectation and the definition of $\mathcal{P}_{\theta,\eta,t}^{\pm}$, the contrastive component satisfies
\begin{equation}
\begin{aligned}
    &\mathbb{E}\!\left[
    \widehat{\mathcal{R}}^{\mathrm{ctr}}_t(\eta,\theta)
    \mid \tilde{B}\neq\emptyset
    \right] \\
    &\quad =
    \mathbb{E}_{(x,Y)\sim\mathcal{P}_{\theta,\eta,t}^{\pm}}
    [\ell_{\mathrm{R}}(\eta;x,Y,t)].
\end{aligned}
\end{equation}
The value and KL terms are ordinary minibatch estimates of their corresponding population expectations. Thus, non-empty minibatches provide Monte Carlo estimates of the population objective components, while empty contrastive minibatches correspond to null contrastive updates rather than unbiased estimates of the contrastive risk.

\textbf{Outer descent (reward model update).} In Step~2 of \cref{alg:co_training}, the reward head and backbone are updated by descending $\mathcal{L}_R$ (\cref{eq:continual_reward_loss}), while the value head is updated by descending $\mathcal{L}_V$ (\cref{eq:continual_value_loss}). The stop-gradient operators define a locally frozen block surrogate of \cref{eq:rm_game_loss}: during the reward-head update, the anchor is held fixed; during the value-head update, the reward targets are held fixed. With sufficiently small step sizes, each block update decreases its corresponding surrogate objective, giving a first-order block-coordinate approximation to descent on the joint outer risk \cite{wright2015coordinate}. Recomputing the sign partition and anchors at the next iteration then refreshes the surrogate around the updated model.

\textbf{Inner ascent (policy update).} In Step~3, the policy is updated to maximize expected reward under the improved RM, with KL regularization as in \cref{eq:rlhf_objective}. By Lemma~1, under Conditions~(C1)--(C3), this update direction has positive inner product with $\nabla_\theta\mathcal{M}_t(\eta,\theta)$. Thus, the policy update is an approximate ascent step on the inner objective, in the sense that it shifts the response distribution toward samples that increase the current reward model's self-supervised risk.

The KL term prevents arbitrary distributional drift, keeping the policy in the local region where the alignment argument applies. Alternating the reward model block descent with the policy ascent step yields a stochastic block-coordinate descent-ascent approximation to the minimax problem in \cref{eq:minimax_game}.

\section{Detailed Training Algorithm}
\label{sec:detailed_algorithm}

\cref{alg:co_training_full} provides the full training procedure corresponding to the compact algorithm in the main text. At each iteration, \method first samples grouped on-policy responses and filters out ambiguous samples using the value-anchored RM advantage. The retained responses are partitioned into positive and negative subsets, which provide self-supervised feedback for updating the reward head, while the value head is calibrated to the group mean reward. After the reward model update, the filtered responses are recomputed with the improved RM and used to optimize the policy with a standard critic-free RL algorithm. This expanded version makes the filtering, reward model update, and policy update stages explicit.

\begin{algorithm*}[t]
\caption{Self-Supervised Reward Model Improvement via Value-Anchored On-Policy Feedback}
\label{alg:co_training_full}
\textbf{Input} Instruction set $X$, initial policy model $\pi_{\theta_0}$, initial reward model parameterized by $\omega_0, \phi_0, \psi_0$, initial margin $m_0$, total training steps $T$, number of candidates $G$.
\begin{algorithmic}[1]
\For{training step $t = 1$ \textbf{to} $T$}
    \State Compute $p(t) \gets \min\!\left(1 - m_0 + \tfrac{t}{T},\; 1\right)$, \; dynamic threshold $\mu(t) \gets \tfrac{2 - p(t)}{2}$
    
    \State Sample a batch of instructions $B \subset X$
    
    \Comment{\textit{Step 1: Adaptive Feedback Filtering}}
    \For{each instruction $x_i \in B$}
        \State Sample $G$ candidate responses $Y_i = \{y_{i, 1}, \dots, y_{i, G}\} \sim \pi_{\theta_{t-1}}(\cdot \mid x_i)$
        \State Initialize filtered response set $\tilde{Y}_i \gets \emptyset$
        
        \State Compute RM advantages $a^{\mathrm{RM}}_{i,j} \gets R_{\omega_{t-1}, \phi_{t-1}}(x_i, y_{i,j}) - V_{\omega_{t-1}, \psi_{t-1}}(x_i)$ for all $j \in [G]$
        \State $\tilde{Y}_i \gets \bigl\{y_{i,j} \mid \sigma\!\bigl(\left|a^{\mathrm{RM}}_{i,j}\right|\bigr) \ge \mu(t)\bigr\}$ \Comment{Retain informative samples}
        \State Split $\tilde{Y}_i$ into $\tilde{Y}_i^{+}$ and $\tilde{Y}_i^{-}$ by the sign of each corresponding $a^{\mathrm{RM}}_{i,j}$
    \EndFor
    \State Filter batch $\tilde{B} \gets \{x_i \in B \mid \tilde{Y}_i^{+} \neq \emptyset \wedge \tilde{Y}_i^{-} \neq \emptyset\}$

    \Comment{\textit{Step 2: Self-Supervised Reward Model Improvement}}
    \State Initialize temporary parameters $\omega_t \gets \omega_{t-1}, \phi_t \gets \phi_{t-1}, \psi_t \gets \psi_{t-1}$
    \State Update $\omega_t, \phi_t$ by gradient descent on reward loss $\mathcal{L}_R(\omega_t, \phi_t)$ (\cref{eq:continual_reward_loss,eq:positive_reward_loss,eq:negative_reward_loss})
    \State Update $\psi_t$ by gradient descent on value loss $\mathcal{L}_V(\psi_t)$ (\cref{eq:continual_value_loss})
    
    \Comment{\textit{Step 3: Policy Model Optimization}}
    \State Recompute RM advantages and filtering on the same sampled groups $\{Y_i\}_{x_i \in B}$ using updated RM $(\omega_t, \phi_t, \psi_t)$ to obtain $\{\tilde{Y}_{i, \text{new}}\}_{x_i \in B}$
    \State Compute updated rewards $R_{\text{new}} \gets \{ R_{\omega_t, \phi_t}(x_i, y) \mid x_i \in B, y \in \tilde{Y}_{i, \text{new}} \}$ using parameters $\omega_t, \phi_t$
    \State Update policy $\pi_{\theta_{t-1}} \to \pi_{\theta_t}$ on the batch $B$ with filtered sets $\{\tilde{Y}_{i, \text{new}}\}_{x_i \in B}$ and rewards $R_{\text{new}}$ using a standard RL algorithm
\EndFor
\end{algorithmic}
\textbf{Output} Improved reward model $(\omega_T,\phi_T,\psi_T)$ and optimized policy model $\pi_{\theta_T}$
\end{algorithm*}

\section{Experiments}

\subsection{Training Dynamics}
\label{sec:training_dynamics}

\begin{figure}[t]
    \centering
    \begin{subfigure}[b]{\linewidth}
        \centering
        \includegraphics[width=\linewidth]{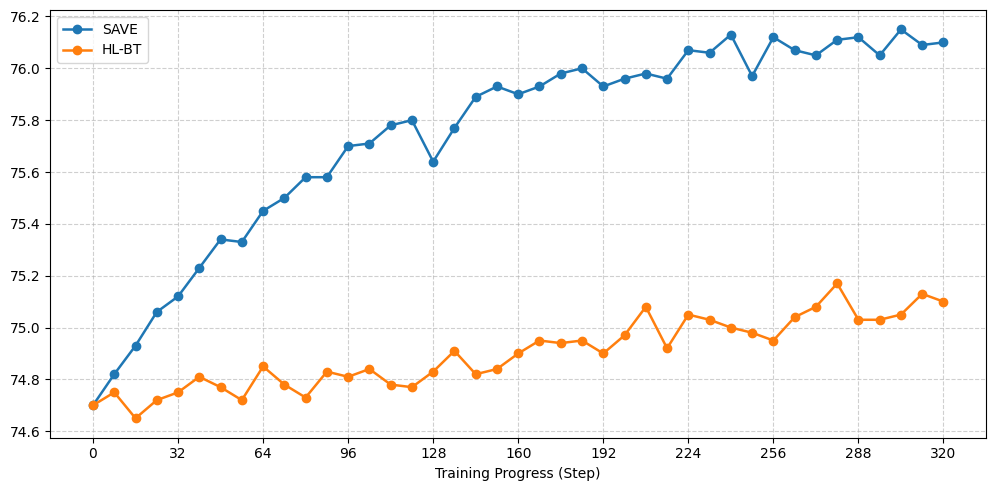}
        \caption{RewardBench 2}
        \label{fig:train_rb2}
    \end{subfigure}
    \\
    \begin{subfigure}[b]{\linewidth}
        \centering
        \includegraphics[width=\linewidth]{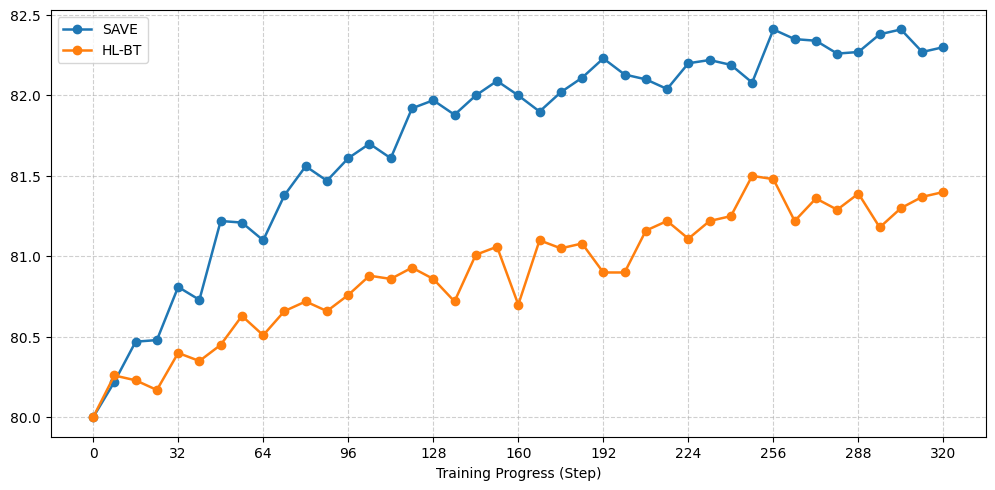}
        \caption{RM-Bench}
        \label{fig:train_rmbench}
    \end{subfigure}
    \caption{Training dynamics of \method compared with baselines over training steps on RewardBench 2 and RM-Bench. All experiments use Qwen3-4B-Instruct-2507 as the policy model with GRPO.}
    \label{fig:training_dynamics}
\end{figure}

\cref{fig:training_dynamics} shows that both \method and HL-BT improve as training progresses on RewardBench 2 and RM-Bench, yet \method rises noticeably faster in the early stages and maintains a consistent lead throughout training. In particular, HL-BT improves slowly and begins to plateau after around 200 steps, while \method continues to climb and converges at a higher level.

\subsection{Sensitivity to Learning Rates}
\label{sec:lr_ablation}

\cref{tab:lr_ablation} examines the sensitivity of \method to the learning rates of the backbone $f_\omega$ \& reward head $h_\phi$ and the value head $v_\psi$. Overall, \method is relatively stable across the tested learning rate combinations, but the best configuration uses a smaller learning rate for the shared backbone and reward head, together with a larger learning rate for the value head. This behavior is consistent with the roles of the two components. The backbone and reward head encode the RM's preference knowledge, so overly aggressive updates may disturb the pretrained reward landscape and amplify noisy self-supervised signals. In contrast, the value head serves as a lightweight prompt-specific anchor and needs to adapt quickly to the current on-policy response distribution. Separating the update scales allows the value anchor to track distributional changes while preserving the reward model's semantic judgment ability.

\subsection{Effect of Group Size}
\label{sec:group_size_ablation}

\cref{tab:group_size_ablation} examines the effect of the number of candidate responses $G$ sampled per prompt. Larger groups generally lead to better reward model performance because they expose the RM to a richer set of on-policy responses under the same instruction. This makes the prompt-specific value anchor more informative and gives the adaptive filter a clearer basis for separating high-confidence positive and negative feedback from near-anchor ambiguous samples. With too few responses, the sampled group may not cover enough quality variation, making the derived RM advantages noisier and less useful for contrastive learning. The improvements become moderate as $G$ grows, suggesting a trade-off between feedback quality and sampling cost; we therefore use $G=16$ as a practical setting in the main experiments.

\subsection{Sensitivity to Initial Margin}
\label{sec:margin_ablation}

\cref{tab:margin_ablation} examines the sensitivity of \method to the initial margin $m_0$, which controls the strictness of the adaptive feedback filtering mechanism in \cref{eq:response_filter}. The margin determines the precision--coverage trade-off of self-supervised feedback. A smaller margin admits more samples, but it also increases the chance of training on responses whose RM advantages are close to the value anchor and therefore unreliable. A larger margin filters more aggressively, improving confidence in the retained samples but discarding potentially useful training signals, especially when the policy distribution is still evolving. The intermediate setting provides the best overall balance, indicating that \method benefits from filtering ambiguous feedback without making the contrastive update overly sparse.

\begin{table*}[t]
    \centering
    \small
    \begin{tabular}{p{0.2\textwidth} p{0.8\textwidth}}
        \toprule
        \textbf{Instruction} &
        Please answer this: Determine the topic of the passage. ``Halley's Comet last appeared in the inner Solar System in 1986 and will next appear in mid-2061.'' Topic: \newline
        ++++++++ \newline
        Answer: Halley's Comet \newline\newline
        Problem: Determine the topic of the passage. ``Bleeding Kansas, Bloody Kansas or the Border War, was a series of violent political confrontations involving anti-slavery Free-Staters and pro-slavery ``Border Ruffian'' elements, that took place in the Kansas Territory and the neighboring towns of Missouri between 1854 and 1861.'' Topic: \newline\newline
        A: Bleeding Kansas \newline\newline
        Problem: Given the question: Determine the topic of the passage. ``A gristmill (also: grist mill, corn mill or flour mill) grinds grain into flour.'' Topic: \newline
        +++++++++++++++++++++++++++++++++ \newline
        The answer is: \newline
        Gristmill \newline\newline
        input question: Determine the topic of the passage. ``A recurring antagonist, he is the Supreme Commander of the Confederacy of Independent Systems, a political faction of planetary systems waging war on the Galactic Republic.'' Topic:??? \newline
        output answer: General Grievous \newline\newline
        Determine the topic of the passage. ``The attack on Pearl Harbor (called Hawaii Operation or Operation AI by the Japanese Imperial General Headquarters (Operation Z in planning) and the Battle of Pearl Harbor) was a surprise military strike conducted by the Imperial Japanese Navy against the United States naval base at Pearl Harbor, Hawaii, on the morning of December 7, 1941 (December 8 in Japan).'' Topic: \newline
        {-}{-}{-}{-} \newline
        Answer: Attack on Pearl Harbor \newline\newline
        Q: Determine the topic of the passage. ``The first color cinematography was by means of additive color systems such as the one patented in England by Edward Raymond Turner in 1899 and tested in 1902.'' Topic: \newline
        A: \\
        \midrule
        \textbf{Response 1} & Color cinematography \\
        \textbf{Response 2} & Color motion-picture photography \\
        \midrule
        \textbf{Reward of Response 1} & 7.469 \\
        \textbf{Reward of Response 2} & 7.719 \\
        \textbf{Value of Instruction} & 5.375 \\
        \bottomrule
    \end{tabular}
    \caption{Case study illustrating how HL-BT overfits to surface-form differences between semantically equivalent responses. Both answers are correct, yet HL-BT treats them as a preference pair and forces the RM to enlarge the reward gap, while \method assigns both to the positive subset $\tilde{Y}^{+}$ via value-anchored partitioning and avoids spurious training signals.}
    \label{tab:case_study}
\end{table*}

\begin{table}[t]
    \centering
    \renewcommand{\arraystretch}{1.3}
    \resizebox{\linewidth}{!}{
    \begin{tabular}{ll c}
        \toprule
        \textbf{Stage} & \textbf{Hyperparameter} & \textbf{Value} \\
        \midrule
        \multirow{5}{*}{\makecell[l]{Value Anchor \\ Integration}}
        & Num.\ responses per prompt $K$ & 16 \\
        & Value head learning rate & $2 \times 10^{-4}$ \\
        & Batch size & 32 \\
        & Training epochs & 1 \\
        \midrule
        \multirow{12}{*}{\makecell[l]{On-Policy \\ Feedback \& \\ RL Training}}
        & Total training steps $T$ & 320 \\
        & Batch size & 32 \\
        & Group size $G$ & 16 \\
        & Initial margin $m_0$ & 0.5 \\
        & Value loss weight $\lambda_V$ & 1 \\
        & KL penalty coefficient $\beta$ & 0.04 \\
        & Backbone \& reward head learning rate & $1 \times 10^{-7}$ \\
        & Value head learning rate & $1 \times 10^{-5}$ \\
        & Policy model learning rate & $2 \times 10^{-6}$ \\
        & Max generation length & 1024 \\
        & Temperature & 0.7 \\
        & Top-$p$ & 0.8 \\
        & Top-$k$ & 20 \\
        \bottomrule
    \end{tabular}}
    \caption{Hyperparameter settings for \method.}
    \label{tab:hyperparams}
\end{table}

\section{Case Study}

\cref{tab:case_study} illustrates a representative failure mode of the HL-BT baseline. Both ``Color cinematography'' and ``Color motion-picture photography'' are semantically equivalent correct answers that differ only in surface form. Because HL-BT pairs the highest and lowest reward responses within each on-policy group and trains with the Bradley-Terry loss, it treats this pair as a chosen/rejected preference signal and forces the reward model to enlarge the gap between them ($7.719$ vs.\ $7.469$). Since no genuine quality difference exists, the reward model can only exploit superficial cues such as phrasing style or lexical choice, gradually overfitting to structural patterns rather than true semantic quality and becoming increasingly susceptible to reward hacking. In contrast, \method avoids this degenerate pairing through value-anchored partitioning. Both rewards ($7.469$ and $7.719$) lie above the value anchor ($5.375$), yielding positive RM advantages for both responses. Consequently, both responses are assigned to the positive subset $\tilde{Y}^{+}$ and placed on the same side of the contrastive objective in \cref{eq:positive_reward_loss}, rather than being pitted against each other across the chosen/rejected divide. This prevents the reward model from being trained on spurious preference pairs and preserves its focus on meaningful quality distinctions.

\section{Implementation Detials}

\subsection{More Experiment Setting}
\label{sec:more_training_details}
To improve training stability, we select UltraFeedback instructions whose responses exceed 1024 tokens. For continual offline training, we use the corresponding instruction-response pairs from \texttt{ultrafeedback\_binarized}, matched to the same instruction set.

During the value anchor integration stage, we first generate $K = 16$ responses for each instruction using Qwen2.5-3B-Instruct and Qwen3-4B-Instruct-2507, respectively. We then use the corresponding generated responses to train the value head $v_\psi$, resulting in two distinct value heads, each fitted to one policy model.

During the reward model update, we further filter out responses longer than 1024 tokens to improve training stability. This prevents excessively long responses from introducing unstable reward estimates, truncation artifacts, or disproportionately large gradient contributions during RM optimization.

\begin{figure}
    \centering
    \includegraphics[width=0.6\linewidth]{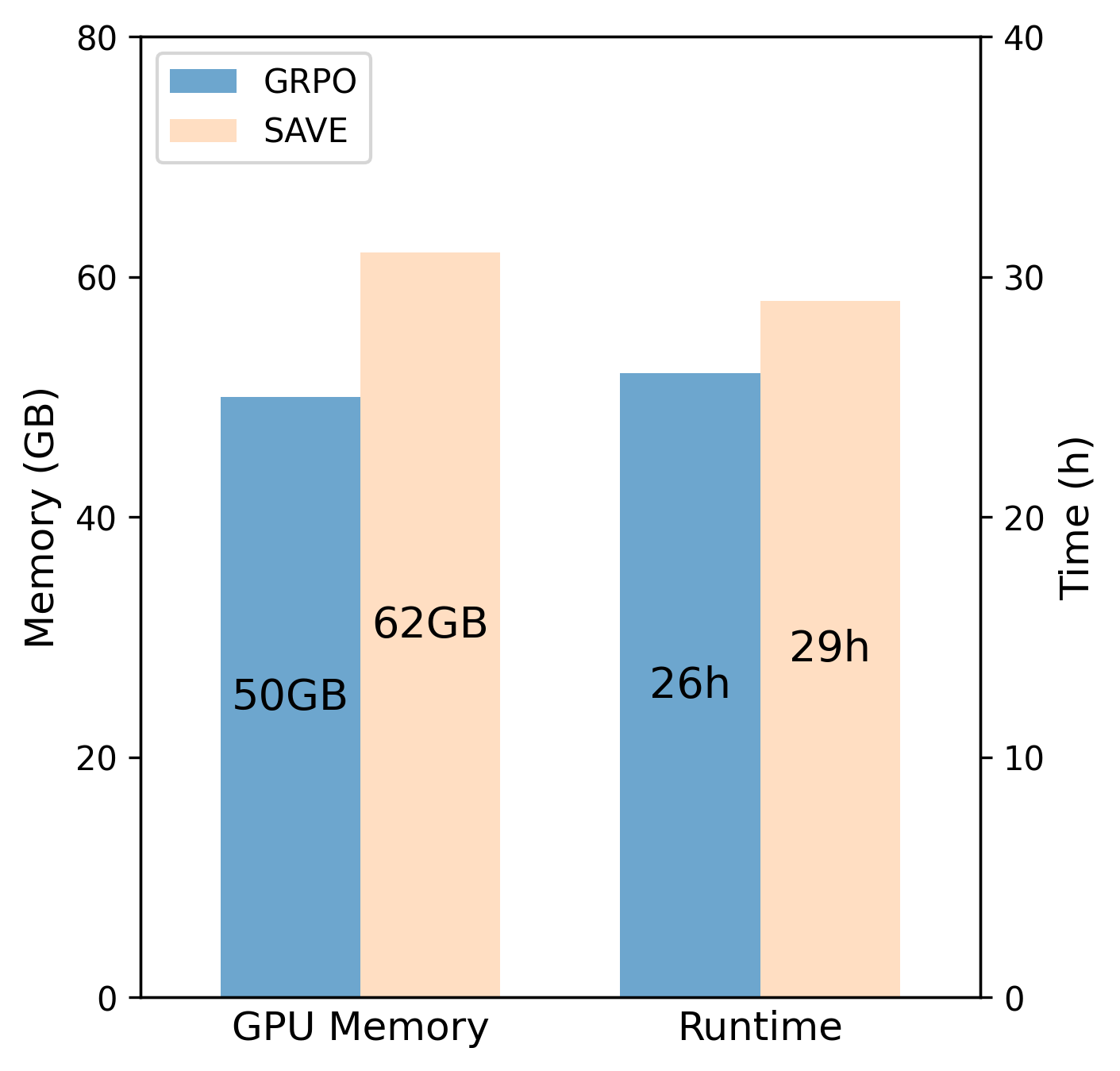}
    \caption{Training cost comparison between GRPO and \method on Qwen3-4B-Instruct-2507 with DeepSpeed ZeRO-2 offload. The left group reports peak GPU memory during the backward pass, and the right group reports total training time under the same experimental setting.}
    \label{fig:memory_runtime}
\end{figure}

\subsection{Hyperparameter Settings}
\label{sec:hyperparams}

\cref{tab:hyperparams} lists the hyperparameters used in our experiments, covering the key settings for data sampling, reward model, and value head optimization, policy learning, and response generation. Unless otherwise specified, we keep these settings fixed across datasets and backbone models to ensure a consistent comparison among different training variants.

\subsection{Computing Resources}
All experiments are conducted on a machine with four NVIDIA A100 80GB GPUs, 32GB system memory, and a 128-core AMD CPU. All training runs use CUDA 12.8, PyTorch 2.7.1, and DeepSpeed 0.18.2.

\subsection{Training Cost Analysis}
\cref{fig:memory_runtime} compares the training cost of standard GRPO and \method under the same experimental setting, both using DeepSpeed ZeRO-2 with CPU offloading. GRPO requires 50GB of peak GPU memory during the backward pass and 26 hours of training, while \method requires 62GB of peak GPU memory and 29 hours. The additional memory mainly comes from maintaining and updating the reward model, including the value head. By comparison, the increase in time is limited: although \method introduces an extra reward model improvement step, it reuses the sampled on-policy responses and only increases total runtime by 3 hours. This suggests that \method trades a moderate increase in memory for stronger reward feedback, while keeping the overall training time comparable to the GRPO baseline.

\section{AI Usage Statement}

We use AI for language polishing, grammar checking, and improving the clarity and conciseness of the manuscript.

\end{document}